\documentclass[preprint, 12pt]{elsarticle}

\usepackage{amssymb}
\usepackage{amsmath}
\usepackage{subcaption}

\usepackage{url}
\usepackage{hyperref}

\begin{document}

\begin{frontmatter}



\title{Hybrid-Segmentor: A Hybrid Approach to Automated Fine-Grained Crack Segmentation in Civil Infrastructure}

\author{June Moh Goo\corref{cor1}\fnref{label1}}
\ead{june.goo.21@ucl.ac.uk}

% \ead[url]{www.linkedin.com/in/jmgoo1118}
% \fntext[label1]{}
\cortext[cor1]{}
\affiliation[label1]{organization={Department of Civil, Environmental and Geomatic Engineering, University College London},
            addressline={Gower Street},
            city={London},
            postcode={WC1E 6BT},
            country={United Kingdom}}
% \author{June Moh Goo} %% Author name
\author[label3]{Xenios Milidonis}
\author[label3]{Alessandro Artusi}
\author[label1]{Jan Boehm}
\author[label2]{Carlo Ciliberto}

% \affiliation[label1]{organization={Department of Civil, Environmental and Geomatic Engineering, University College London},%Department and Organization
%             addressline={Gower Street}, 
%             city={London},
%             postcode={WC1E 6BT}, 
%             % state={},
%             country={United Kingdom}}
\affiliation[label3]{organization={DeepCamera MRG, CYENS Centre of Excellence},
            % addressline={Gower Street}, 
            city={Nicosia},
            % postcode={WC1E 6BT}, 
            country={Cyprus}}
            
\affiliation[label2]{organization={Department of Computer Science, University College London},%Department and Organization
            addressline={Gower Street}, 
            city={London},
            postcode={WC1E 6BT}, 
            country={United Kingdom}}

%% Abstract
\begin{abstract}
Detecting and segmenting cracks in infrastructure, such as roads and buildings, is crucial for safety and cost-effective maintenance. In spite of the potential of deep learning, there are challenges in achieving precise results and handling diverse crack types. With the proposed dataset and model, we aim to enhance crack detection and infrastructure maintenance. We introduce Hybrid-Segmentor, an encoder-decoder based approach that is capable of extracting both fine-grained local and global crack features. This allows the model to improve its generalization capabilities in distinguish various type of shapes, surfaces and sizes of cracks.
To keep the computational performances low for practical purposes, while maintaining the high the generalization capabilities of the model, we incorporate a self-attention model at the encoder level, while reducing the complexity of the decoder component.
The proposed model outperforms existing benchmark models across 5 quantitative metrics (accuracy 0.971, precision 0.804, recall 0.744, F1-score 0.770, and IoU score 0.630), achieving state-of-the-art status.
\end{abstract}

% % %%Graphical abstract
%     \includegraphics[width=\textwidth]{graphical_abstract.pdf}

% % %%Research highlights
% \begin{highlights}
% \item Introduced Hybrid-Segmentor, combining CNN and Transformer paths for enhanced crack segmentation.
% \item Experiments on a refined dataset of 12,000 images
% \item Achieved state-of-the-art performance on 5 metrics: Accuracy, Precision, Recall, F-1 Score and IoU.
% \item Model generalizes well across diverse crack shapes and surfaces
% \item Outperforms benchmarks in handling discontinuities, small regions, and low-quality crack contours.
% \end{highlights}

%% Keywords

\end{frontmatter}



\section{Introduction}
\label{intro}
%The detection of structural damage in civil infrastructure is crucial in preventing potential hazards and safeguarding lives while facilitating timely maintenance interventions. 
Cracks in roads, pavements, and buildings pose a serious threat to public safety, causing accidents and damage to vehicles on roads and pavements, and influencing public safety and financial burden on buildings. Traditionally, manual inspections have been used to identify cracks in civil infrastructure, but these methods are labor-intensive, subjective, and prone to human error, resulting in inconsistent results and potential disasters. Therefore, automated crack detection is necessary to provide an objective and highly accurate alternative. Machine learning methods, such as deep learning models, can be used to detect, segment, or classify damage to civil infrastructure, which can be facilitated by the widespread deployment of surveillance and traffic-monitoring cameras. However, training accurate models for crack segmentation is challenging due to a scarcity of well-annotated and diverse datasets, which impacts model robustness and generalizability. Our research aims to address this crucial data gap and develop automated crack detection to prevent dangers and reduce financial risks to communities. Progress in this direction could lead to real-time identification of %newly developed 
cracks in the future, ensuring a more dependable and safe utilization of critical concrete structures. 

%In this paper, we address two main goals: first, we introduce a large dataset for crack segmentation. This effort is aimed at filling a critical gap in the literature on crack detection and ultimately providing a valuable resource to the research community on the topic. Second, we propose a new deep learning architecture, empirically proving its effectiveness on real data. 
The main contributions of this work are:

\begin{itemize}
    \item Combine and refine publicly available crack datasets to create an enhanced and comprehensive crack segmentation dataset.
    \item Introduce a data refinement methodology to combine publicly available datasets using image processing techniques. 
    \item Introduce the Hybrid-Segmentor model to efficiently detect cracks in infrastructures, which is based on the encoder-decoder architecture that convolutional neural networks (CNNs) and transformers have efficiently used in the past.%Develop and evaluate a deep learning-based crack detection model utilizing the merits of both convolutional neural networks and transformers.
    \item Emphasize the remarkable ability of the proposed model to perform effectively across a diverse range of surface types and under challenging imaging conditions, such as blurred images and areas with complex crack contours.
    \item The code, trained weights of the model, and the full dataset for experiments are publicly available and can be accessed here: \\
   \href{https://github.com/junegoo94/Hybrid-Segmentor}{https://github.com/junegoo94/Hybrid-Segmentor}
\end{itemize}

\section{Related Work}
\label{lr}

\begin{figure}[t]
  \centering
  \begin{subfigure}[t]{0.49\linewidth}
    \centering
    \includegraphics[width=\linewidth]{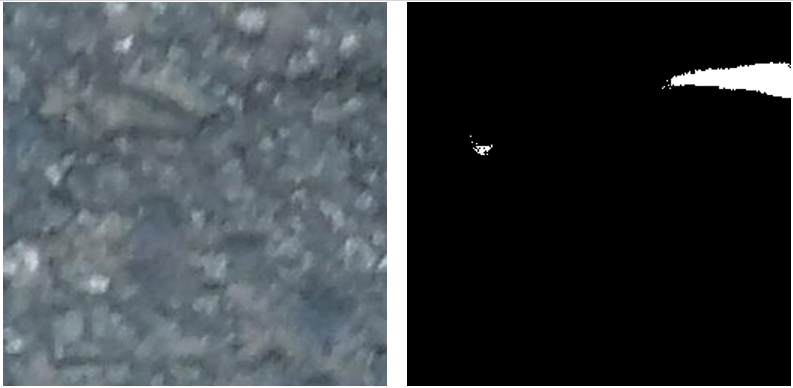}
    \caption{Blurred Image}
    \label{fig:wrong_example_A}
  \end{subfigure}
  \hfill
  \begin{subfigure}[t]{0.49\linewidth}
    \centering
    \includegraphics[width=\linewidth]{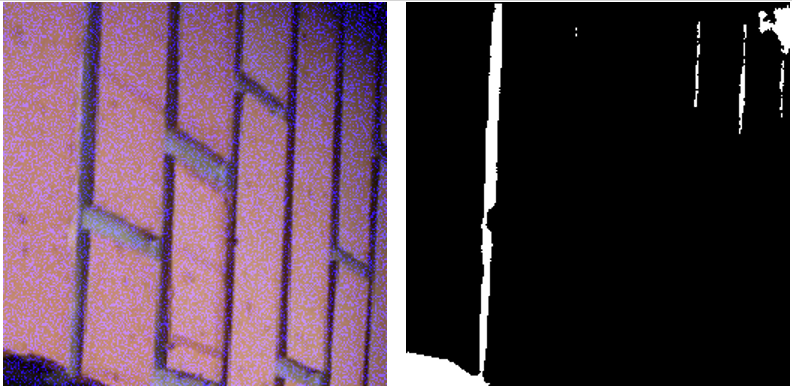}
    \caption{Brick Image}
    \label{fig:wrong_example_B}
  \end{subfigure}
    \caption{Representative failures in crack detection by traditional models: (a) shows a prediction by a Fully Convolutional Network (FCN) on a blurred image, incorrectly marked as a crack, highlighting difficulties with image clarity. (b), from a U-Net architecture, displays a brick pattern where the borders of the bricks are wrongly identified as cracks, revealing challenges in differentiating structural boundaries}
  \label{fig:wrong_examples}
\end{figure}

One of the earliest methods for crack detection was a CNN-based model for pixel-level crack detection using FCN \cite{crack_FCN}. This approach achieves end-to-end crack detection, significantly reducing training time compared to CrackNet \cite{cracknet}, a CNN-based model that was the State-Of-The-Art (SOTA) in 2017 without using a pooling layer. While thin cracks can be accurately predicted across a variety of scenes, further enhancements are needed to capture real-time level predictions. In a similar aspect, DeepCrack \cite{deepcrack_dataset} improves on the generalization of FCN architecture by incorporating batch normalization and side networks for faster convergence. Additionally, this research proposes the publicly available DeepCrack dataset \cite{deepcrack_dataset}, which enhances crack detection precision across diverse scenes. Cheng et al. propose a full crack segmentation model based on U-Net  \cite{crack_unet}. Subsequent research further demonstrated that the U-Net is particularly suited for crack segmentation tasks \cite{crack_unet, attention_unet, crack_ImageNet_unet, UWnet, panella2022semantic}. Some researchers pinpoint that using classical image classification structures as encoders, pre-trained with data such as ImageNet \cite{imagenet}, strengthens feature extraction in crack segmentation networks, enhancing crack detection performance \cite{crack_ImageNet_unet}. 

In addition, various encoder-decoder models have been introduced in the field. Amongst these, DeepCrack2 (not to be confused with `DeepCrack' in \cite{deepcrack_dataset} bearing the same name; we refer to this model as `DeepCrack2' from this point onwards to avoid confusion) is a deep convolutional neural network designed to facilitate automated crack detection through end-to-end training \cite{crack_deepcrack2}. It primarily focuses on acquiring high-level features that effectively represent cracks. This approach involves the integration of multi-scale deep convolutional features obtained from hierarchical convolutional stages. This fusion enables the capture of intricate line structures, with finer-grained objects in larger-scale feature maps and more holistic representations in smaller-scale feature maps. DeepCrack2 adopts an encoder-decoder architecture similar to SegNet \cite{segnet} and employs pairwise feature fusion between the encoder and decoder networks at corresponding scales. DeepCrack2 is one of the most benchmarked models in the crack segmentation community.

Despite the abundance of studies that either employ existing deep learning models or enhance them, these approaches may not always result in effective or efficient results in real-world scenarios (Fig.\ref{fig:wrong_examples}). Recently, HrSegNet \cite{crack_hrsegnet} was proposed as an approach to consistently maintain high resolution in the images, distinguishing itself from methods that restore high-resolution features from low-resolution ones. Furthermore, the model enhances contextual information by leveraging low-resolution semantic features to guide the reconstruction of high-resolution features \cite{crack_hrsegnet}. These features helped HrSegNet-B64 reach SOTA accuracy and inference speed in crack segmentation.

\section{Dataset}
We introduce a large refined dataset with the aim of creating a significantly larger and more diverse resource for crack segmentation compared to what is currently available in literature. Since the existing datasets contain a relatively small number of images compared to other well-known tasks in computer vision, large-scale deep learning models are at a high risk of overfitting in these settings. In contrast to most datasets for crack segmentation that collect data based on a single type of surface, the refined comprehensive dataset includes a wide range of surfaces to enhance the robustness and generalizability of trained models. %by adapting to the diversity of the real world. 
Additionally, due to the characteristics of some cracks, each existing image has a small proportion of crack pixels, which could result in a form of class imbalance. To counteract this bias, we employed a data augmentation strategy to increase the number of crack pixels in our dataset. 

\subsection{Sub-Dataset Details}
We identified 13 open-source datasets that include different surfaces of pavements, walls, stone, and bricks. Table \ref{table:raw_dataset_details} shows the details of each dataset. 
Some datasets provide samples either collected with specific acquisition systems and under diverse background settings (e.g. Aigle-RN, ESAR, and LCMS that collectively form the AEL Dataset) \cite{AEL_dataset}; or acquired with smartphone cameras (e.g. CRACK500) \cite{crack500_dataset}. A number of small datasets provide road and pavements images, including CrackTree260, CRKWH100, CrackLS315 and Stone331 \cite{Deep_crack_crackLS315}. (e.g. CrackTree260 is a dataset of 260 visible-light road pavement images constructed based on the CrackTree206 \cite{cracktree206})
%Aigle-RN, ESAR, and LCMS (AEL Dataset) are the datasets collected by their own acquisition system, which comprises a small number of asphalt crack images acquired in diverse background settings \cite{AEL_dataset}. CRACK500 is a pavement crack dataset with 500 images captured on Temple University's main campus using smartphones. Each image has a pixel-level annotated binary map. They crop each image into 16 non-overlapping regions \cite{crack500_dataset}. CrackTree260, CRKWH100, CrackLS315 and Stone331 datasets are collected by \cite{Deep_crack_crackLS315}. CrackTree260 with 260 visible-light road pavement images which is constructed based on the CrackTree206 \cite{cracktree206}, CRKWH100 containing 100 road pavement images captured by a line-array camera at a 1 mm ground sampling distance, CrackLS315 consisting of 315 road pavement images captured under laser illumination using the same line-array camera specifications, and Stone331 comprising visible-light images of stone surfaces along with masks for precise performance evaluation within the stone surface area \cite{Deep_crack_crackLS315}. 
DeepCrack \cite{deepcrack_dataset} is a large dataset created as a publicly available benchmark dataset consisting of crack images captured across various scales and scenes, specifically designed to evaluate the performance of crack detection systems. The German Asphalt Pavement Distress (GAPs) dataset, introduced in \cite{GAPS_data_original}, addresses the comparability issue in pavement distress research, offering a standardised dataset with 1,969 high-quality gray valued images. It covers various distress classes, including cracks, potholes, and inlaid patches. The images have a resolution of 1,920 × 1,080 pixels with a per-pixel resolution of 1.2 mm × 1.2 mm. To enable pixel-wise crack prediction, 384 images are manually selected from GAPs and annotated, forming the GAPs384 dataset \cite{FPHBN_gaps384}. Masonry is created consisting of images captured from masonry structures, which exhibit intricate backgrounds and a diverse range of crack types and sizes \cite{masonry_dataset}. CrackForest dataset (CFD), one of the most benchmarked datasets, is a labeled collection of road crack images, designed to represent the typical conditions of urban road surfaces \cite{CFD1,CFD2}. Finally, SDNET2018 is a dataset comprising more than 56,000 images of cracked and non-cracked concrete bridge decks, walls, and pavements, with crack widths ranging from 0.06 mm to 25 mm. Since the dataset does not contain ground truth masks, we use this dataset only for the collection of non-cracked image data \cite{sdnet2018}.

\begin{table}
\caption{Sub-datasets details before data refinement.}
\centering
\resizebox{\linewidth}{!}{%
\begin{tabular}{c|c|c|c|c} 
\hline\hline
\textbf{Dataset} & \textbf{Size}        & \textbf{Resolution} & \textbf{Surface}     & \begin{tabular}[c]{@{}c@{}}\textbf{Crack}\\\textbf{Proportion (\%)}\end{tabular}  \\ 
\hline\hline
Aigle-RN         & 38                   & Various Sizes        & Pavement             & 0.71                                                                              \\
 CFD              & 118                  & 480 x 320           & Pavement             &1.62                                                                              \\
 CRACK500         & 500                  & 2000 x 1500         & Pavement             &6.01                                                                              \\
 CrackLS315       & 315                  & 512 x 512           & Pavement             &0.25                                                                              \\
 CrackTree260     & 260                  & Various Sizes        & Pavement             &0.46                                                                              \\
 CRKWH100         & 100                  & 512 x 512           & Pavement             &0.36                                                                              \\
 DeepCrack        & 537                  & 544 x 388           & Diverse surfaces     &3.5                                                                               \\
 ESAR             & 15                   & 512 x 768           & Pavement             &0.6                                                                               \\
 GAPs384          & 384                  & 640 x 540           & Pavement             &0.36                                                                              \\
 LCMS             & 5                    & 1000 x 700          & Pavement             &0.67                                                                              \\
 Masonry          & 240                  & 224 x 224           & Bricks/Masonry walls &4.21                                                                              \\
 SDNET2018        & 56092 & 256 x 256           & Pavement             &-                                                                                 \\
 Stone331         & 331                  & 512 x 512           & Stone Surfaces       &0.11                                                                              \\ 
\hline\hline
Total Dataset    &                      &                     &                      & 2.69                                                                              \\
\hline\hline
\end{tabular}
}
% \caption{Sub Datasets details before image processing and data-preprocessing.}
\label{table:raw_dataset_details}
\end{table}

\subsection{Data Refinement}
\begin{figure}[t]
  \centering
    \includegraphics[width=\linewidth]{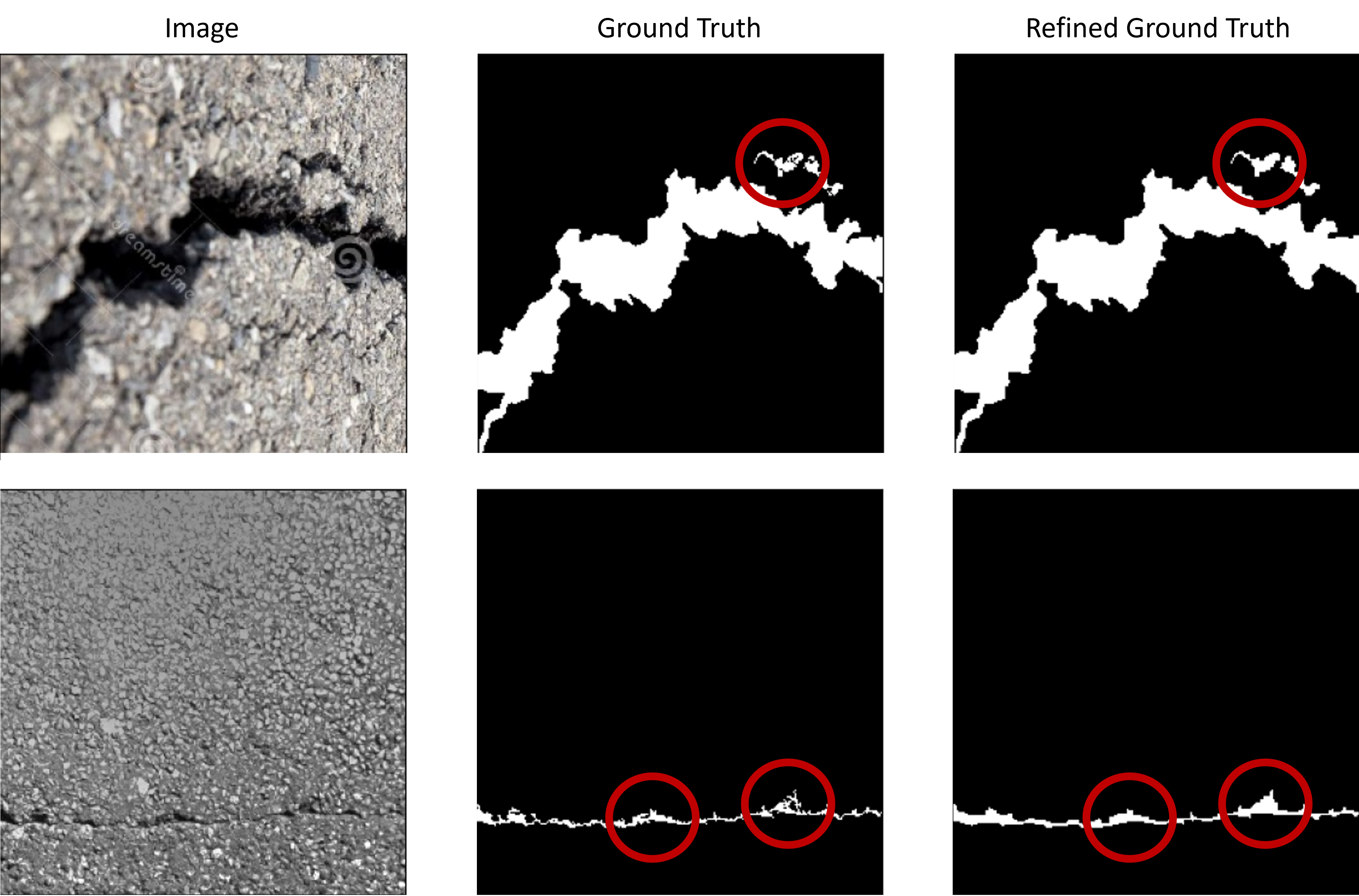}
    \caption{The figure shows the improvement in small holes, discontinuity and thinness of the ground truth after applying appropriate image processing methods.}
    \label{fig:refinement}
\end{figure}

Ground truth masks in existing datasets were generated using different methods, leading to varying resolutions, distortions, and discontinuity. To address this inconsistency, masks were manually inspected and refined using basic image processing where deemed necessary to ensure no irregularities were present, based on a process described previously \cite{crackseg9k}. Due to the inconsistency of AEL datasets with the rest of the datasets (inverted and not binary), dedicated processing steps were performed. First, the values in the masks were inverted. Pixels were then converted to either black or white based on a threshold of 255/2. All images included in our dataset were then cropped to 256 $\times$ 256 resolution without overlapping. Finally, due to the reduced number of images with cracks, we augmented our dataset with a significant portion of cracks to address class imbalance. Specifically, images with masks containing over 5000 crack pixels were selected for augmentation, where Gaussian noise was added, and a random rotation of 90$^\circ$, 180$^\circ$, or 270$^\circ$ was applied. Non-crack data from the SDNet2018 dataset \cite{sdnet2018} were also added.
% Table \ref{table:refined_dataset_details} shows further details of the data refinement strategy and the final dataset.
Fig. \ref{fig:refinement} shows how the original ground truth improved after the refinement process. Irregularities such as small holes, discontinuity, and thinness were corrected. Furthermore, adding the augmented dataset increased the proportion of the crack pixels by 5.8$\%$, which aims to mitigate class imbalance problems. As a result, we created a comprehensive refined dataset with a total of 12,000 images, which is the largest crack dataset to the best of our knowledge.

\section{Model Design}
\begin{figure*}[t]
    \centering
    \includegraphics[width=\textwidth]{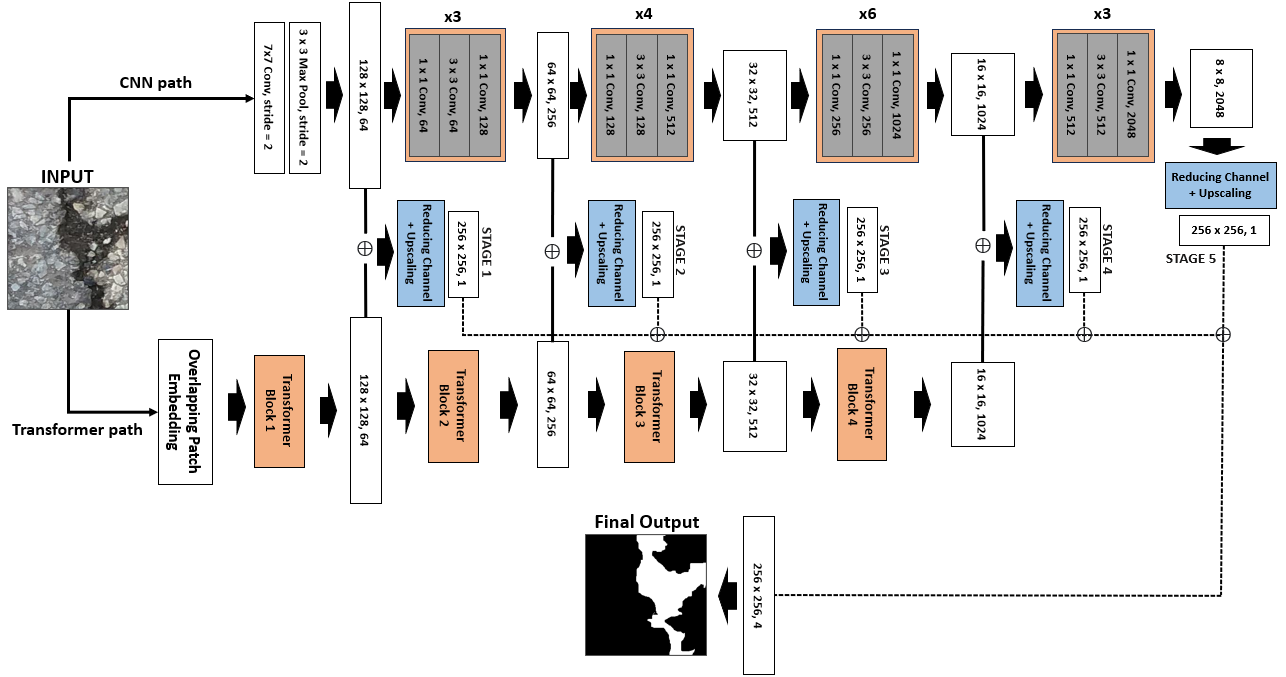}
    \caption{The Hybrid-Segmentor architecture: the upper path for CNN and the lower for Transformers. Each path generates feature maps at every layer, and the central blue boxes represent the concatenation of these feature maps.}
    \label{fig:model_architecture}
\end{figure*}

This section provides an in-depth overview of our Hybrid-Segmentor, an end-to-end crack segmentation model. As illustrated in Fig. \ref{fig:model_architecture}, the input images to our model are going through two distinct encoders: the CNN (ResNet-50 \cite{resnet}) and the Transformers (SegFormer \cite{segformer}) paths. Each of these encoders generates 5 multi-scale feature maps, which are then fused together at each of the 5 intermediate stages. In the last step, the fused feature maps are utilized to produce the final output (simplified decoder).
The overall benefits of our Hybrid-Segmentor, through the combination of the two different deep learning architectures are the ability to detect local details and global structural understanding, while spatial hierarchy leads to more accurate crack detection. Integrating features at different scales from both paths enables effective recognition of cracks of various sizes and shapes, leveraging the strengths of both local and global analysis.
This ensures higher accuracy and robustness in detecting cracks in diverse types of surfaces.
Sections \ref{sec:CNN} and \ref{sec:Transformer} further describe the benefits introduced by the CNN and transformer paths of our architecture respectively.

\subsection{CNN Path}
\label{sec:CNN}
The use of a CNN architecture is guided from the fact that we would like to capture local features from the input image. These features are both fine-grained local details, e.g., small cracks or textures, and high level features, such as abstract shapes. This is achieved through the spatial hierarchy property of the ResNet-50 model used in our Hybrid-Segmentor, which allows the detection of various image features at multiple scales. Additionally, its translation invariant property will help with extracting features regardless of the crack position within the input image. Finally, its capability to preserve high-resolution details will make our model more effective at detecting small cracks or local variations.

\subsection{Transformer Path}
\label{sec:Transformer}
The use of a transformer in crack segmentation aims to extract global features from the input image, which are crucial for capturing the overall shape and appearance of a crack. Here, we use as our base key concepts from the SegFormer model \cite{segformer}. Through its self-attention mechanism, this model can recognise the continuity and structure of cracks that span distant regions, understanding how different parts of the crack relate to each other across the image (Long-range Dependency Capture). SegFormer also incorporates a spatial hierarchy, similar to CNNs, by processing features at different scales, making it able to capture both fine details and global structures. Another important property of the SegFormer is its global consistency, which by analysing the image entirely, it provides insights into how cracks are distributed across the entire image, ensuring a coherent understanding of the crack patterns. 

The transformer of our proposed model utilizes three additional key concepts, which are explained below: Overlapping Patch Embedding \ref{sec:Embedding}, Efficient Self-Attention\ref{sec:Self-Attention}, and Mix-Feed Forward Network (FFN)\ref{sec:FFN}.

\subsubsection{Overlapping Patch Embedding}
\label{sec:Embedding}
Local continuity is crucial for preserving fine-grained details and spatial coherence, which is important for accurate semantic segmentation. The first iterations of vision transformers used non-overlapping patch embeddings, which could lead to a loss of local continuity between patches. However, to address this, we utilized overlapping patch embedding, as introduced by SegFormer \cite{segformer}, which better preserves local continuity.

The Vision Transformer (ViT) \cite{ViT} is an innovative approach to computer vision. It treats images as sequences of patches and processes them similarly to how transformers handle sequences of words in natural language processing. In a typical ViT architecture, an image is divided into N × N patches, which are then linearly embedded into 1 × 1 × C vectors. While this method enables the model to effectively capture global context, it can still be challenging to maintain local continuity among patches when N × N × 3 image patches are represented as 1 × 1 × C vectors.
% It can be challenging to maintain local continuity among patches when $N \times N \times 3$ image patches are represented as $1 \times 1 \times C$ vectors in the Vision Transformer (ViT) \cite{ViT} architecture. 

To address this issue, 
%the Swin Transformer \cite{swin_trans} introduced a Shifted Window approach, which allows for better local continuity by slightly shifting the windows between layers. On the other hand, 
SegFormer employed Overlapping Patch Embedding. Instead of simply dividing the image into non-overlapping $4\times4$ patches for vector embedding, Overlapping Patch Embedding takes inspiration from how CNNs use sliding windows with defined parameters such as kernel size (K), stride (S), and padding (P). It predefines these parameters to split the input image into patches of size $B \times C \times K^2 \times N$, where $B$ represents the batch size, $C$ is the number of channels times the stride squared, and $N$ is the number of patches. Merging operations are then performed to transform the reshaped patches to $B \times C \times W \times H$, where $W$ and $H$ represent the width and height of the merged patches, respectively. As a result, the model captures both fine-grained local details and broader global features more effectively, addressing the issue of losing local continuity while still maintaining global context.

\subsubsection{Efficient Self-Attention}
\label{sec:Self-Attention}
Especially in models like SegFormer with smaller patch sizes like $4 \times 4$, the self-attention layer presents computational challenges. The traditional multi-head attention process involves creating matrices for query (Q), key (K), and value (V), all of which have dimensions $N (H \times W) \times C$, and performing computations using the scaled dot-product attention equation as shown in equation \ref{eq:ESA_1}.
\begin{equation}
    \text{Attention(Q, K, V)} = \text{Softmax}(\frac{QK^T}{\sqrt{d_{head}}})V
\label{eq:ESA_1}
\end{equation}
When dealing with large input images, the computational complexity of the provided equation \ref{eq:ESA_1} can lead to a significant increase in model weight. Therefore, the method that reduces the $N (H \times W)$ channels of $K$ and $V$ by applying a sequence reduction process based on a predefined reduction ratio is proposed \cite{segformer}. It is possible to reshape the equation by dividing $N$ by $R$ and multiplying $C$ by $R$. $C \times R$ dimensions can be reduced to $C$ dimensions by linear operation, resulting in $\frac{N}{R} \times C$ dimensions for Key and Value matrices. Especially useful for tasks like semantic segmentation, this method efficiently manages computational complexity while preserving representation power (equation \ref{eq:ESA_2}).
\begin{equation}
\begin{array}{cl}
    \hat{K} = Reshape(\frac{N}{R}, C \cdot R)(K)\\
    K = Linear(C \cdot R, C)(\hat{K})
\end{array}
\label{eq:ESA_2}
\end{equation}

\subsubsection{Mix-FFN}
\label{sec:FFN}
ViT \cite{ViT} uses positional encoding for local information, which comes with fixed input resolution constraints and suffers performance drops as resolution changes. In order to overcome this issue, researchers replace positional encoding with a Convolutional $3 \times 3$ kernel in the FFN, asserting its non-essential role and providing flexibility without resolution restrictions.
\begin{equation}
    \textbf{x}_{out} = MLP(GELU(Conv_{3\times3}(MLP(\textbf{x}_{in})))) + \textbf{x}_{in}
\label{eq:mix_ffn}
\end{equation}

In this regard, the equation \ref{eq:mix_ffn} simply adds a Convolutional $3\times3$ layer to the existing FFN within the Transformer encoder. By replacing traditional positional encoding with this adaptation, the model performance is maintained while fewer parameters are required.

\subsection{Decoder}
The CNN and transformer paths of our model result in substantial model complexity and size. In order to balance this, the decoder is designed to be as simple as possible. A $256 \times 256 \times 1$ feature map is generated by concatenating the outputs from both paths, as shown in Fig. \ref{fig:model_architecture}. Feature maps from each stage are combined to create a multi-scale, multi-layer feature map, which is then used to create the final output. By integrating the strengths of both paths, this approach optimizes performance and efficiency while simplifying the decoder.

\section{Experimental Settings}
\subsection{Training Setup}
Models were trained and tested with a batch size of 16 on a GPU cluster with 8 nodes, each with 8 NVIDIA RTX A5000 (24 GB on-board GPU memory), running Rocky Linux 8.5 and using Python 3.10 and PyTorch 1.13. For all models, we use early stopping with patience of 10 epochs to ensure convergence and avoid over-fitting.

\subsection{Data}
The refined dataset contains 12,000 images with and without cracks, along with the ground truth for each image. A random shuffling method was used to distribute the dataset between training, testing, and validation sets, with a ratio of 8:1:1. As a result, our dataset consists of 9,600 samples for the training, 1,200 samples for the testing, and 1,200 samples for the validation.

\section{Experiments}
\subsection{Benchmarks}
To assess the performance improvement of our model over traditional segmentation models, we compare it against FCN \cite{FCN} and UNet \cite{UNet}. Additionally, we include the DeepCrack2 model \cite{crack_deepcrack2}, a widely benchmarked crack detection model, as well as SegFormer \cite{segformer} and HrSegNet-B64 \cite{crack_hrsegnet}, which represent SOTA models in semantic segmentation and crack segmentation, respectively. The performance of our model is assessed both quantitatively and qualitatively.

For all models, we used the Adam optimizer, set the initial learning rate to 1.00e-04, and used a batch size of 16. Other hyperparameters are provided in Table \ref{hyperparams_table}. 
\begin{table}[h]
\caption{Hyperparameter settings of benchmarked models (LR indicates Learning Rate)}
\centering
% \resizebox{\linewidth}{!}{%
\begin{tabular}{c|c|c} 
\hline\hline
Model          & LR Schedule       & Pre-trained                                                     \\ 
\hline\hline
Hybrid-Segmentor & ReduceLROnPlateau & \begin{tabular}[c]{@{}c@{}}ResNet \\(IMAGENET1K)\end{tabular}   \\ 
\hline
HrSegNet       & ReduceLROnPlateau & None                                                            \\ 
\hline
DeepCrack2     & None              & None                                                            \\ 
\hline
SegFormer      & PolynomialLR      & None                                                            \\ 
\hline
UNet           & None              & None                                                            \\ 
\hline
FCN            & None              & \begin{tabular}[c]{@{}c@{}}VGG19\\(IMAGENET1K)\end{tabular} \\
\hline\hline
\end{tabular}
% }
% \caption{Hyperparameter setting of benchmarked models (LR refers to Learning Rate)}
\label{hyperparams_table}
\end{table}

\subsection{Loss Functions}
Experimentally, it has been demonstrated that class imbalance in datasets may be effectively addressed not only using a well designed architecture, but also using a well designed loss function \cite{NGUYEN2023116988, sudre2017generalised, recall_loss}. To improve the robustness of our model, we evaluated the performance of various loss functions: Binary Cross Entropy (BCE) \cite{BCE}, Dice \cite{sudre2017generalised}, the fusion of BCE and Dice \cite{NGUYEN2023116988}, and Recall Cross Entropy (RecallCE) \cite{recall_loss}.

The BCE loss has been chosen for its ability to handle 
%is widely used for segmentation tasks with class imbalance as it handles 
skewed pixel distributions effectively. In scenarios where one class significantly outweighs others, BCE loss computes an individual loss for each pixel to ensure proportional class contribution, mitigating any dataset bias. By treating pixels equally, the model is able to focus on accurately classifying minority classes, such as crack pixels, without being biased by dominant classes. BCE loss is described by the following equation:

\begin{equation}
   BCE(y, \hat{y}) = -\frac{1}{N}(y_i\: log(\hat{y})+(1-y_i)\, log(1-\hat{y_i}))) 
\end{equation}
where $N$ is the total number of elements (pixels in the case of segmentation). $y_i$ is the ground truth label (0 or 1) for the i-th element, and $\hat{y}_i$ is the predicted probability for the i-th element.

Dice loss, which is equivalent to F1-score, is addressing the class imbalance focusing on capturing the overlap between predicted and ground truth masks, which helps address the challenge of minority class representation. By emphasizing object boundaries and assigning non-vanishing gradients to the minority class, Dice loss ensures accurate prediction and better learning for smaller classes. 

\begin{equation}
    \text{Dice Loss} = 1 - \frac{2 \cdot Intersection}{Union}
\end{equation}

Its ability to sensitively measure the similarity between prediction and ground truth makes it particularly useful for precise segmentation. It can be used alongside other losses such as BCE loss to keep a balance between handling class imbalance and capturing fine details \cite{NGUYEN2023116988}. Here, we used a combination of BCE and Dice losses as follows:
\begin{equation}
    \text{BCE-DICE} = \lambda * \text{BCE loss} + (1-\lambda) * \text{Dice loss}
\end{equation}
where $\lambda$ represents the weight (importance) attributed to the two loss functions and takes values between 0 and 1.

Previous methods attempt to improve standard cross-entropy loss in segmentation tasks by incorporating weighted factors. However, this approach can lead to issues such as reduced precision and increased false positive rate for minority classes. To address this problem, RecallCE loss is proposed as a hard-class mining solution. It reshapes the traditional cross-entropy loss by dynamically adjusting class-specific loss weights according to a real-time recall score, offering a more effective way to handle class imbalance and improve segmentation precision \cite{recall_loss}. We evaluate the performance of our model by comparing the RecallCE loss with the other losses previously mentioned, to determine if it enhances our model's effectiveness. The equation for RecallCE loss is as follows:
\begin{equation}
    \text{RecallCE} = -\sum_{c=1}^{C}\sum_{n:y_i=c}^{}(1-R_{c,t})\:log(p_{n,t})
\end{equation}
where $R_{c,t}$ represents the recall of class $c$ during optimisation iteration $t$.

\section{Evaluation}
We carry out two prior studies to examine specific aspects of our model: (1) assessing the impact of individual encoder paths and (2) evaluating the performance of various loss functions. Initially, we aim to understand how distinctively each encoder extracts features. Then, we investigate which of the aforementioned individual loss functions and the combination of BCE and Dice losses (by assigning different weights) yields the best results in crack segmentation. Once we determine the loss function providing the optimal performance, we compare our final model against SOTA crack segmentation models.

\subsection{Encoder Paths}
We conduct an experiment involving the training and testing of the two different encoders to assess their abilities in feature extraction. Specifically, we aim to determine whether convolutional layers perform well at extracting local features while transformers are adept at capturing global features. Each path was trained as an independent network by removing the influence of the other, and was compared against the fused network. An identical loss function was used for all networks for a fair comparison (BCE-DICE loss with $\lambda$ = 0.5).

\begin{figure}[]
  \centering
    \includegraphics[width=\linewidth]{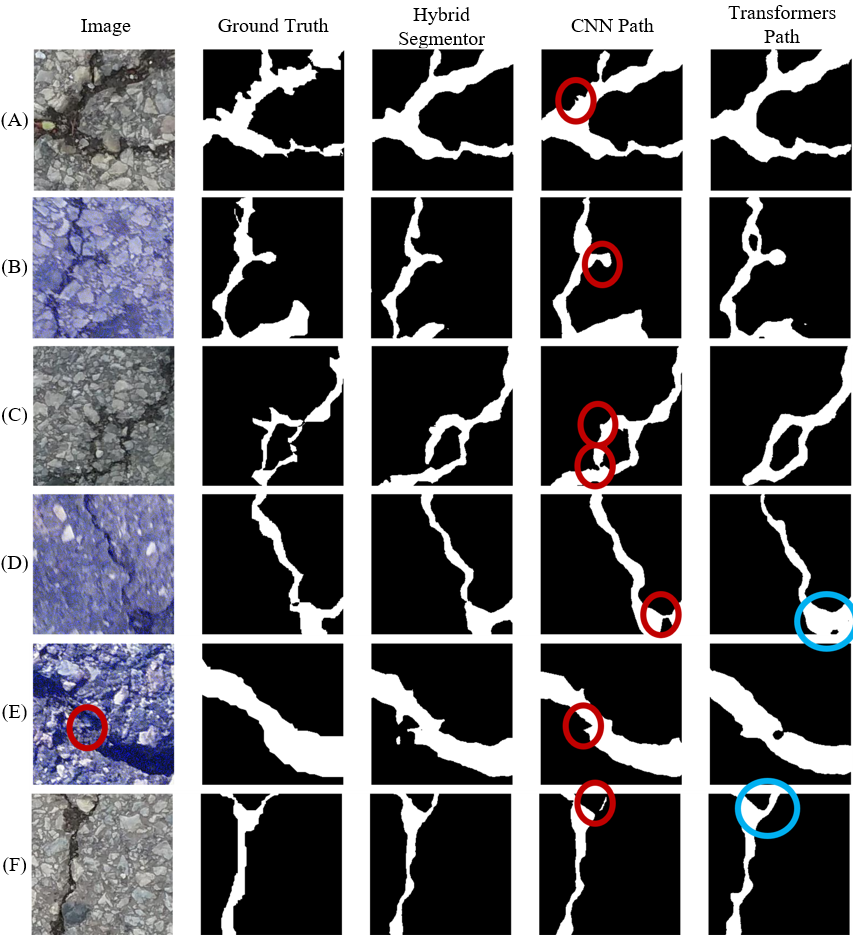}
    \caption{Hybrid model performance compared against CNN and transformer paths. The CNN path captures detailed contours (red circles), while the Transformer path gives an overall structure but with thicker predictions (blue circles). }
    \label{fig:qual_encoder_comparison}
\end{figure}

The results presented in Table \ref{table:encoder_results} indicate that the CNN path achieves a higher precision score than the transformer path, while the latter excels in terms of recall. This suggests that the transformer tends to produce more false positives, mistakenly predicting non-crack pixels as cracks. On the other hand, the CNN path tends to produce more false negatives, possibly misclassifying crack pixels as non-cracks. These results suggest that the transformer path captures broader areas as cracks, while the CNN path captures finer details. Combining the two paths into a fused model leverages the power of both and improves the accuracy and precision of crack segmentation, without significantly sacrificing recall. 
Fig. \ref{fig:qual_encoder_comparison}, shows example segmentations produced by each of the two encoders, further illustrating the differences in their performance. 
% The CNN path excels in capturing intricate contours of the cracks (Red-circled area), while the Transformer path provides an overview of the crack shapes, resulting in slightly thicker predictions compared to the ground truth (Blue-circled area).

\begin{table}[b]
\caption{Comparison and Performance Analysis of CNN and Transformer Paths}

\centering
\resizebox{\linewidth}{!}{%
\begin{tabular}{c|c|c|c|c|c} 
\hline\hline
\textbf{Model Name}& \textbf{Accuracy} & \textbf{Precision}                & \textbf{Recall}                   & \begin{tabular}[c]{@{}c@{}}\textbf{F1 Score}\\\textbf{(Dice)}\end{tabular} & \textbf{IOU score}  \\ 
\hline\hline
\textbf{Hybrid-Segmentor (combined)}& \textbf{0.970}& \textbf{0.805}& \textbf{0.732}& \textbf{0.765}& \textbf{0.622}\\
CNN path& 0.969& 0.802& 0.722& 0.758& 0.614\\
Transformer path& 0.965& 0.717& 0.772& 0.741& 0.592\\
\hline\hline
\end{tabular}
}
% \caption{Quantitative analysis on both CNN and Transformer paths}
\label{table:encoder_results}
\end{table}

\subsection{Loss functions}

We utilize various types of losses (BCE, Dice, and RecallCE) for addressing class imbalances and capturing fine-grained details. Our experiments reveal that combining BCE and Dice losses provides a balance between recognizing dominant classes and accurately segmenting minority groups, resulting in a more effective model for imbalanced data than when using the loss functions individually (Table \ref{table:loss_functions_results}). We assess these aspects by varying the weights assigned to BCE and DICE loss functions. When BCE and DICE loss weights are roughly equal, the model generally performs better. A BCE-DICE loss with $\lambda$ = 0.2 outperforms other values in all metrics except for precision.  Precision peaks at 0.817, but this trades off with recall, resulting in relatively lower performance in other metrics.
Expectedly, RecallCE results in the highest recall score, as it penalizes the model heavily for false negatives while producing well-balanced results for the other metrics. However, this loss is behind BCE-DICE in terms of accuracy and precision, indicating that it may be less effective at addressing class imbalance.
In summary, the BCE-DICE loss with $\lambda$ = 0.2 exhibits the best model performance, and was chosen as the loss function for our final model. 
% Please refer to Table \ref{table:model_comparison} comparing the performance of our final model with other benchmarked models.

\begin{table}[t]
\caption{Performance of all combinations of loss functions}
\centering
\resizebox{\linewidth}{!}{%
\begin{tabular}{c|c|c|c|c|c} 
\hline\hline
\begin{tabular}[c]{@{}c@{}}\textbf{Loss Functions}\\\textbf{($\lambda$)}\end{tabular} & \textbf{Accuracy}                 & \textbf{Precision}                & \textbf{Recall}                   & \begin{tabular}[c]{@{}c@{}}\textbf{F1 Score}\\\textbf{(Dice)}\end{tabular} & \textbf{IOU score}                 \\ 
\hline\hline
DICE                                                                           & 0.970& 0.807& 0.727& 0.763& 0.620\\
BCE-DICE (0.1)                                                                  & 0.970& 0.796& 0.741& 0.765& 0.622\\
BCE-DICE (0.2)                                                                  & \textbf{0.971}& 0.804& 0.744& \textbf{0.770}& \textbf{0.630}\\
BCE-DICE (0.3)                                                                  & 0.970& 0.809& 0.719& 0.759& 0.615\\
BCE-DICE (0.4)                                                                  & 0.970& 0.809& 0.720& 0.760& 0.616\\
BCE-DICE (0.5)                                                                  & 0.970& 0.805& 0.732& 0.765& 0.622\\
BCE-DICE (0.6)                                                                  & 0.970& 0.805& 0.736& 0.767& 0.625\\
BCE-DICE (0.7)                                                                  & 0.970& 0.808& 0.724& 0.762& 0.618\\
BCE-DICE (0.8)                                                                  & 0.969& \textbf{0.817}& 0.700& 0.752& 0.605\\
BCE-DICE (0.9)                                                                  & 0.969& 0.804& 0.719& 0.757& 0.612\\
BCE                                                                            & 0.969& 0.778& 0.750& 0.762& 0.618\\
RecallCE                                                                    & 0.970& 0.795& \textbf{0.746}& 0.768& 0.626\\
\hline\hline
\end{tabular}
}
% \caption{Quantitative analysis on all combinations of loss functions}
\label{table:loss_functions_results}
\end{table}

\subsection{Comparison against SOTA models}
We compare our best model using BCE-DICE loss ($\lambda = 0.2$) to the benchmark models in our experiment. As demonstrated in Table \ref{table:model_comparison}, our model notably outperforms the other five models. Our model achieved an accuracy of 0.971, a precision of 0.804, a recall of 0.744, an F1-score of 0.770, and an IOU score of 0.630. These results demonstrate the model's exceptional proficiency in crack segmentation tasks.

\begin{table}[t]
\caption{Performance of our crack segmentation model against state-of-the-art models}
\centering
\resizebox{\linewidth}{!}{%
\begin{tabular}{c|c|c|c|c|c} 
\hline\hline
\textbf{Model Name}& \textbf{Accuracy}                 & \textbf{Precision}                & \textbf{Recall}                   & \begin{tabular}[c]{@{}c@{}}\textbf{F1 Score}\\\textbf{(Dice)}\end{tabular} & \textbf{IOU score}                 \\ 
\hline\hline
FCN                 & 0.968& 0.802& 0.702& 0.746& 0.598\\
UNet                & 0.968& 0.789& 0.720& 0.750& 0.603\\
DeepCrack2           & 0.968& 0.788& 0.704& 0.741& 0.592\\
SegFormer           & 0.965& 0.750& 0.719& 0.730& 0.580\\
HrSegNet            & 0.969& 0.800& 0.724& 0.757& 0.612\\
 Hybrid-Segmentor& \textbf{0.971}& \textbf{0.804}& \textbf{0.744}& \textbf{0.770}&\textbf{0.630}\\
 \hline \hline
\end{tabular}
}
% \caption{Quantitative analysis on our proposed model and benchmarked models}
\label{table:model_comparison}
\end{table}

\begin{figure}[!t]
  \centering
    \includegraphics[width=\textwidth]{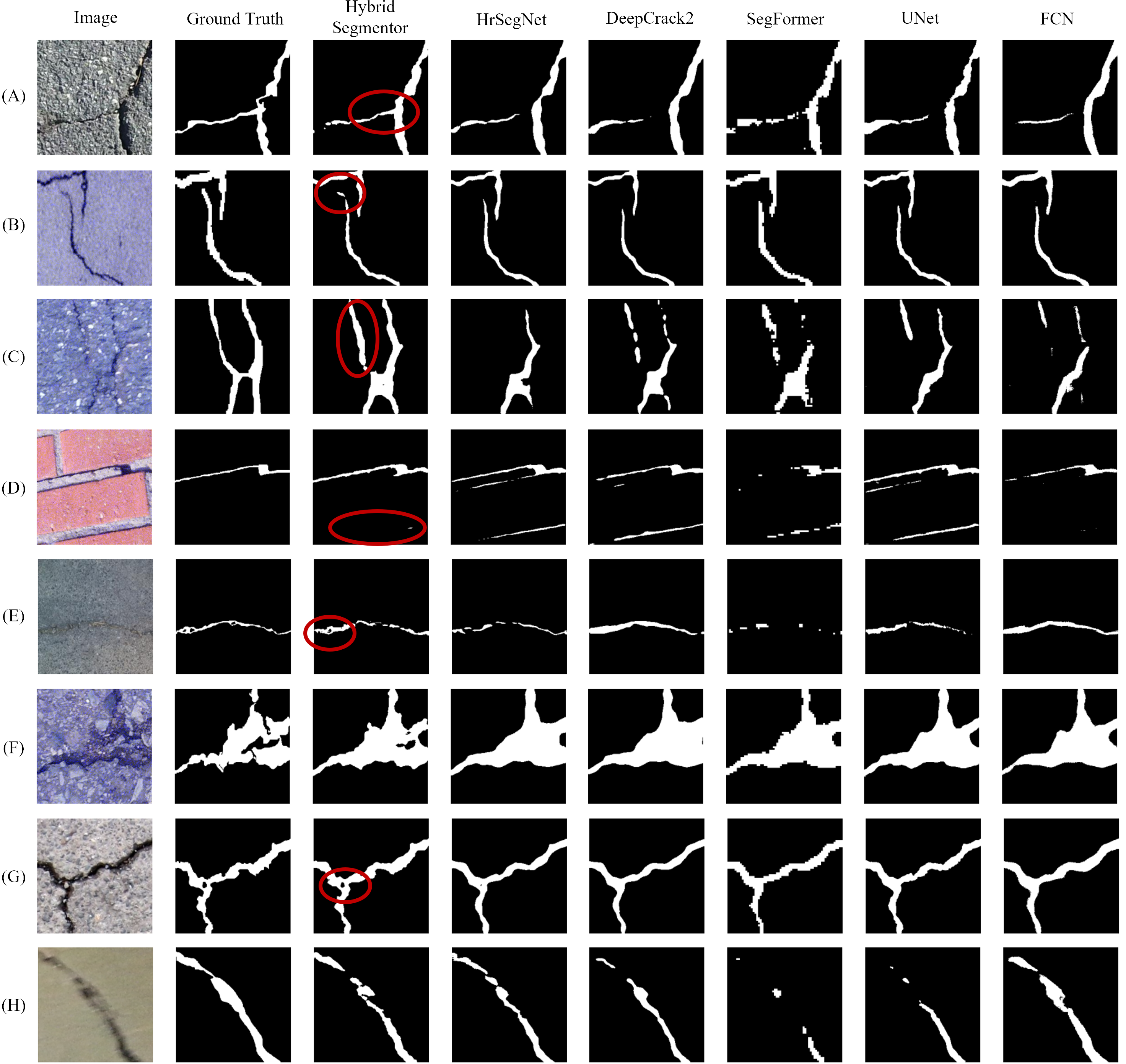}
    \caption{Example crack images segmented by our model and benchmarked models. The red ovals highlight the areas where our model outperforms other benchmarked models. In examples without red ovals, such as (F) and (H), our model demonstrates strong performance across overall structures. }
    \label{fig:qual_model_comparison}
\end{figure}

Qualitatively, our model exhibits significant improvements relative to existing models (Fig. \ref{fig:qual_model_comparison}). As shown by rows (A) and (C), our model handles crack discontinuity more accurately. Furthermore, in (B), our model excels at identifying vague cracks that other models fail to detect. When it comes to cracks on different types of surfaces, the proposed model works effectively regardless of the surface. While crack detection on brick surfaces is challenging due to the ambiguity between cracks and brick borders and resulting shadows, as shown in (D), our model is adept at handling such scenarios. On the other hand, models such as FCN incorrectly predict brick borders as cracks. Additionally, a challenge in crack detection involves identifying non-crack areas within cracked regions, which our model effectively addresses, as evident in (E) and (G). Example (H) demonstrates that our model works relatively well on blurred images. Furthermore, (F) demonstrates the superiority of our model in detecting intricate crack contours, in comparison to the other models that have significantly more false positives.

\section{Limitations}
\begin{figure}[!t]
  \centering
    \includegraphics[width=\textwidth]{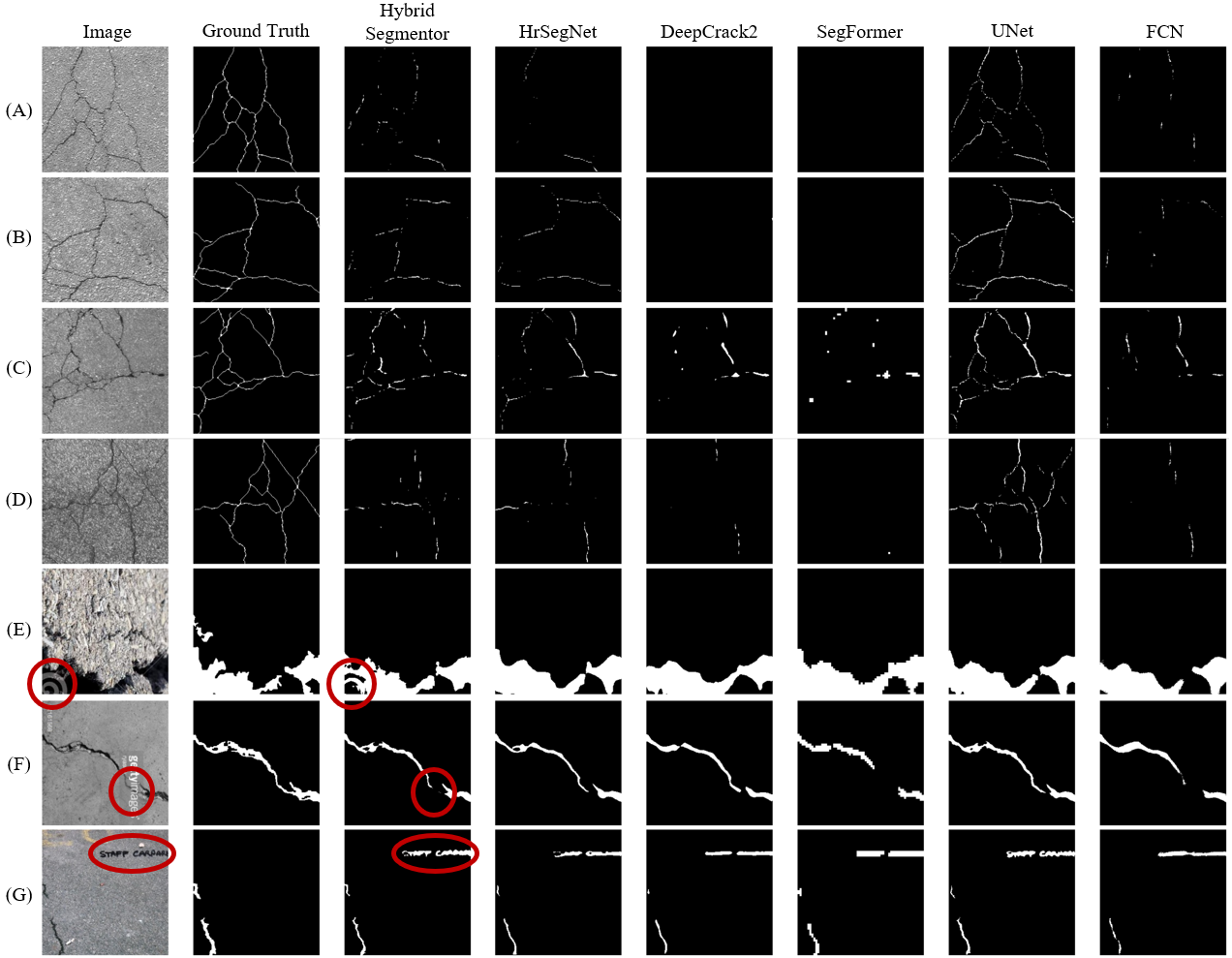}
    \caption{Examples of Hybrid-Segmentor's limitations, including failure to detect thin or web-shaped objects and difficulties with occlusions.}
    \label{fig:limitaitons}
\end{figure}

Although our model outperforms other benchmarked models in performance, it still exhibits certain limitations. Two primary shortcomings of our model have been identified and presented in Fig. \ref{fig:limitaitons}.
% Firstly, our model struggles with the detection of thinner cracks within web-shaped crack patterns. Example images (A) to (D) illustrate web-shaped cracks, for which our model fails to detect the thin branches. while thicker ones are successfully identified.
% 
Our model outperforms the other models in detecting thicker cracks within web-shaped crack patterns, except for UNet; however, it still faces challenges in identifying the thinner branches in these patterns. As illustrated in example images (A) to (D), while our model successfully detects the most prominent cracks, it struggles with extremely fine and delicate ones. This indicates an area where further improvement is possible.

Secondly, our model is sensitive to disruptions caused by distortions, such as occlusions and watermarks. (E) illustrates a situation where the watermark located within the crack is identified as a non-crack area. However, in (F), even with the presence of a watermark, our model fails to predict the cracks hidden by a translucent occlusion. Furthermore, (G) demonstrates an issue where the model does not recognize letters on the road as part of the background. The variation in model performance may be attributed to the clear color contrast between the letters and the background, causing confusion for the model. It should be emphasized that all these challenges are common to all crack detection models. However, our model demonstrates better overall precision and elaboration in crack detection compared to others.

We believe that there is room for improvement to the model architecture for addressing these limitations. Additionally, techniques such as Generative Adversarial Networks (GANs) and meta-learning can be harnessed to generate synthetic data during the pre-processing phase to further deal with the lack of data and potential class imbalance issues. Furthermore, recognizing the increasing need for 3D crack image segmentation, the creation of high-quality 3D crack image datasets becomes imperative to advance this domain.

\section{Conclusion}
In this research, we have proposed a novel model for crack segmentation called Hybrid-Segmentor. This architecture incorporates two distinct encoder paths, namely the CNN path and the Transformer path. For the CNN path, we use the well-established ResNet-50 architecture\cite{resnet}, which is renowned for its ability to extract local features. Additionally, we introduce the concept of Overlapping Patch Embedding, Efficient Self-Attention, and Mix-FFN in the Transformer path, derived from the SegFormer \cite{segformer} model. In combining two encoders, these additions optimize computational efficiency and model size, thereby soothing capacity problems. We further simplify the model with a relatively simpler decoder to minimize its size.

Through experimentation, Hybrid-Segmentor emerges as a SOTA, outperforming other renowned benchmark models. Our model effectively takes advantage of the two encoder paths, as proved by prior studies evaluating its performance in extracting local and global crack features. Based on the findings of previous studies, the BCE-DICE loss, weighted at 0.2 on the BCE loss, yields the best performance. In qualitative analysis, our model improves in addressing discontinuities, detecting small non-cracked areas within cracks, and recognizing cracks even in low-quality images and diverse surfaces. It is capable of capturing more details in crack contours than previous models.

Furthermore, our study introduces a data refinement methodology for combining publicly available datasets comprising 13 open-source crack datasets with refined ground truths. Since these datasets initially used diverse standards for creating ground truth, we merge and improve them to ensure equivalence, thereby increasing their reliability and precision. In addition, we employ a specific data augmentation approach in order to address the issue of class imbalance within our dataset. By extracting data containing cracks with more than 5000 pixels and augmenting them, we are able to incorporate these samples into our dataset. Our effort resulted in a dataset consisting of 12,000 crack images, each with its corresponding ground truth.

To enhance our crack detection model, we need to concentrate on enhancing the architecture to efficiently recognize thin, web-shaped cracks, and those that are hidden by occlusions. We could also explore the possibility of using GANs and meta-learning to create synthetic data to overcome data scarcity, particularly in the development of 3D crack image segmentation datasets.

\section{Acknowledgment}
The research work of Dr. Alessandro Artusi and Dr. Xenios Milidonis has been partially funded from the European Union’s Horizon 2020 research and innovation programme under grant agreement No. 739578 and from the Government of the Republic of Cyprus through the Deputy Ministry of Research, Innovation and Digital Policy.

% \appendix
% \section{Example Appendix Section}


% %% For citations use: 
% %%       \cite{<label>} ==> [1]

% %%
% Example citation, See \cite{lamport94}.





%% For numbered reference style
\bibliographystyle{elsarticle-num.bst}
\bibliography{references.bib}
%   \textit{\LaTeX: a document preparation system},
%   Addison Wesley, Massachusetts,

% \end{thebibliography}
\end{document}